\documentclass[letterpaper, 10 pt, conference]{ieeeconf}

\IEEEoverridecommandlockouts                              

\usepackage{graphicx}
\usepackage{amsmath}
\usepackage{amssymb}
\usepackage{booktabs}
\usepackage{times}
\usepackage{epsfig}
\usepackage{graphicx}
\usepackage{amsmath}
\usepackage{amssymb}
\usepackage{amsfonts}
\usepackage{wrapfig}
\usepackage{cite}
\usepackage{color,soul}
\usepackage[ruled,vlined]{algorithm2e}
\usepackage{lipsum}
\usepackage{cuted}
\usepackage{mathrsfs }
\usepackage{mathtools}
\usepackage{algorithmic}
\usepackage{enumerate}
\usepackage{makeidx}  
\usepackage{bm}
\usepackage{verbatim}
\usepackage{multirow}
\usepackage{float}
\usepackage{booktabs}
\usepackage{textcomp}
\usepackage{url}
\usepackage{arydshln}
\usepackage{color,soul}
\usepackage{subfloat}
\usepackage[caption=false]{subfig} 

\usepackage[capitalize]{cleveref}
\crefname{section}{Sec.}{Secs.}
\Crefname{section}{Section}{Sections}
\Crefname{table}{Table}{Tables}
\crefname{table}{Tab.}{Tabs.}

\DeclareMathOperator*{\argmin}{argmin}

\newtheorem{Def}{Definition}

\begin{document}

\title{\bf DIME-Net: Neural Network-Based Dynamic Intrinsic Parameter Rectification for Cameras with Optical Image Stabilization System}

\author{Shu-Hao~Yeh, Shuangyu~Xie, Di~Wang, Wei~Yan, and Dezhen~Song\\
\thanks{S. Yeh, S. Xie, D. Wang, and D. Song are with Computer Science and Engineering Department, Texas A\&M University, College Station, TX 77843, USA. Emails: \textit{ericex1015@tamu.edu}, \textit{sy.xie@tamu.edu}, \textit{ivanwang@tamu.edu}, and \textit{dzsong@cs.tamu.edu}.}
\thanks{W. Yan is with Architecture Department, Texas A\&M University, College Station, TX 77843, USA. Email: \textit{wyan@tamu.edu}.}
\thanks{S. Yeh and S. Xie are co-first authors of this paper.}
\thanks{This work was supported in part by National Science Foundation under IIS-2119549 and NRI-1925037, and by GM/SAE Autodrive Challenge II.}
}
\maketitle
\thispagestyle{empty}
\pagestyle{empty}

\begin{abstract}
Optical Image Stabilization (OIS) system in mobile devices reduces image blurring by steering lens to compensate for hand jitters. However, OIS changes intrinsic camera parameters (i.e. $\mathrm{K}$ matrix) dynamically which hinders accurate camera pose estimation or 3D reconstruction. Here we propose a novel neural network-based approach that estimates $\mathrm{K}$ matrix in real time so that pose estimation or scene reconstruction can be run at camera native resolution for the highest accuracy on the mobile devices.
Our network design takes gridified projection model discrepancy feature and 3D point positions as inputs and employs a Multi-Layer Perceptron (MLP) to approximate $f_{\mathrm{K}}$ manifold. We also design a unique training scheme for this network by introducing a Back propagated PnP (BPnP) layer so that reprojection error can be adopted as the loss function. The training process utilizes precise calibration patterns for capturing accurate $f_{\mathrm{K}}$ manifold but the trained network can be used anywhere. We name the proposed Dynamic Intrinsic Manifold Estimation network as DIME-Net and have it implemented and tested in three different mobile devices. In all cases, DIME-Net can reduce reprojection error by at least $64\%$ indicating that our design is successful.

\end{abstract}

\section{Introduction}
\begin{figure}[t]
\centering
\subfloat[\label{fig::OIS}]{\includegraphics[height = 1.5in, trim={0.5cm 0.5cm 0.5cm 0.5cm}, clip]{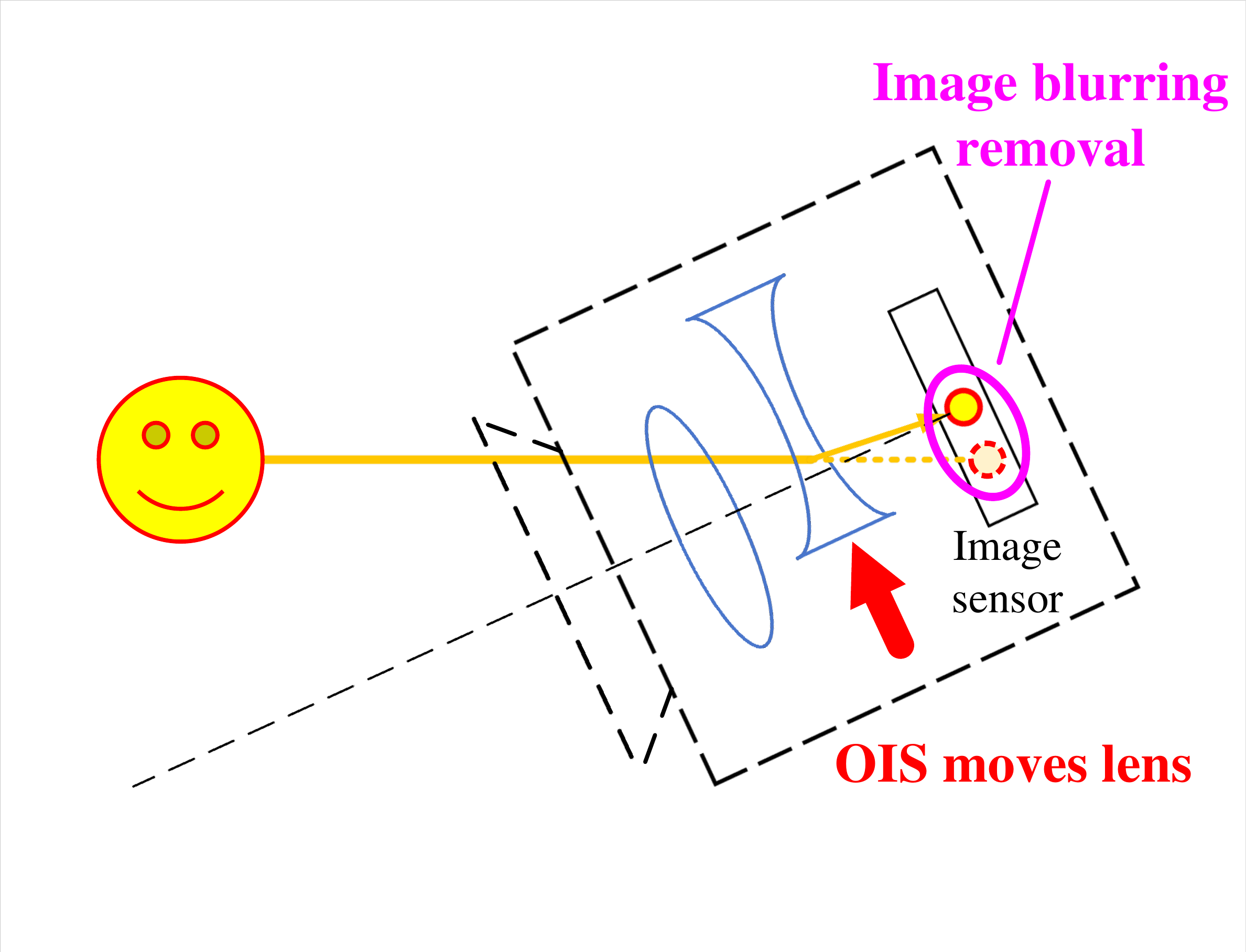}}\hspace*{.2in}\hspace*{-.15in}
\subfloat[\label{fig::system_diagram}] {\includegraphics[height = 1.5in, trim={0.5cm 0.5cm 0.5cm 0.5cm}, clip]{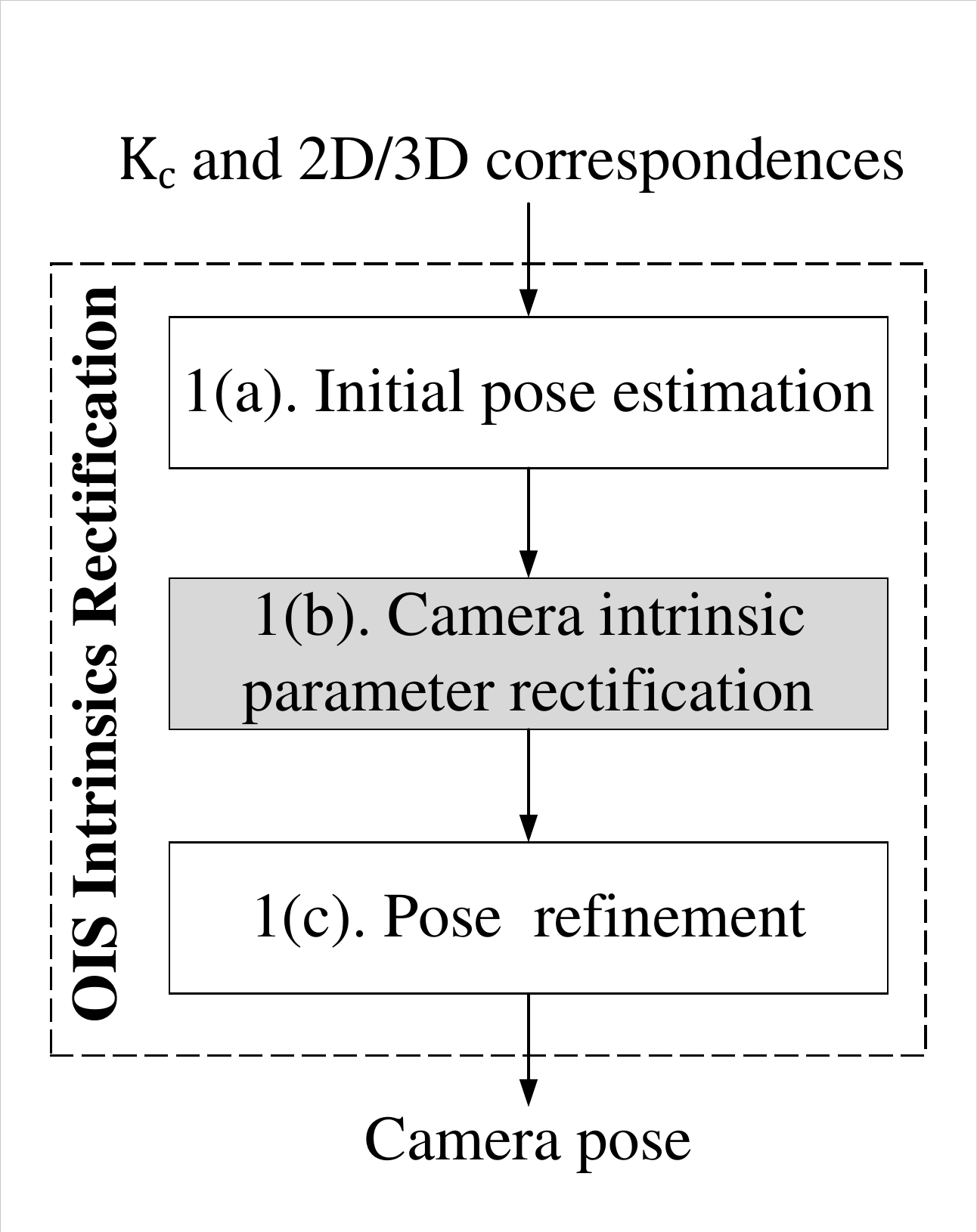}}
\caption{(a) Illustration of OIS working principle. (b) System diagram of OIS intrinsics rectification algorithm for PnP problem.}
\label{fig::problem_background}
\end{figure}

Fig.~\ref{fig::OIS} illustrates how typical OIS functions. When a camera on a mobile device looks at a point in the scene and the camera-holding hand jitters, the image blurs because the point is imaged as a short trajectory instead of a point. To mitigate this effect, the camera is often equipped with a motion sensor to sense the hand/camera motion. The sensed motion is used to generate a countering motion to actuate camera lens or part of lens array so that the imaged point remains at the same 2D location in the imaging sensor and the stationary part of image remains sharp.

Unfortunately, OIS dynamically changes camera intrinsics (i.e. $\mathrm{K}$ matrix) which makes it difficult to accurately estimate camera pose or perform scene reconstruction. Such applications are often seen in visual simultaneous localization and mapping or augment reality which are widely deployed in mobile devices. Existing practice opts to reduce camera resolution so that $\mathrm{K}$ can be approximated by an averaged value which clearly sacrifices accuracy in results.

Here we propose a novel neural network based approach to predict $\mathrm{K}$ matrix. To our best knowledge, this is a new method to solve a new problem since there is no existing work to solve the dynamic $\mathrm{K}$ under OIS effect. To illustrate the network design with real application, we use Perspective-n-Point (PnP) problem~\cite{lepetit2009epnp} for camera pose estimation as an application example. Our network achieves real time estimation for the $\mathrm{K}$ matrix regardless of input image resolution so that downstream tasks, such as camera pose estimation and scene reconstruction algorithms, can run at camera native resolution. For the network input, we propose a gridified project model discrepancy feature and 3D point positions. The network architecture employs an MLP to predict a new $\mathrm{K}$ matrix. We also design a unique training scheme for this network by introducing a Back-propagated PnP (BPnP) layer~\cite{chen2020end} so that reprojection error can be adopted as the loss function. The training can be done using precise calibration patterns in lab settings which builds manifold approximation using carefully collected data to ensure good knowledge embedding. The network inference does not require a large number of high quality features and can be applied to nature objects. We name the Dynamic Intrinsic Manifold Estimation network as DIME-Net and have it implemented and tested in three different mobile devices. DIME-Net can reduce reprojection error by at least $64.0\%$ indicating that our design is successful.

\section{Related Work}

In a nutshell, our approach is to train a neural network to track dynamic intrinsic camera parameters. It is related to camera projection modeling, calibration, and geometry-guided neural network.

Camera projection modeling describes the mapping between a $3$D world coordinate and a $2$D image coordinate. It often consists of extrinsic and intrinsic camera parameters which are often referred to as extrinsics and intrinsics for brevity, respectively. Extrinsics are $6$-Degrees of Freedom (DoFs) camera pose in world coordinate system whereas intrinsics characterize camera and lens internal properties (e.g. focal length, principal point). The perspective projection model~\cite{hartley2003multiple,kannala2006generic}, also known as pinhole model, is the most widely adopted camera model. For a fixed lens camera, its intrinsics are a constant matrix. Of course, this is not true for an OIS-activated camera. Cameras with alterable optical configurations like telephoto lens exist and have variable intrinsics~\cite{lavest1993three,li1996some,simon1999registration,miura2000active,wu2012keeping}. To model this type of camera, control parameters of optical settings become inputs to intrinsic functions. Similarly, camera developers can obtain servo actuator measurement of OIS to estimate intrinsics, such as CIP-VMobile~\cite{jin2020camera}. For regular users, most manufacturers do not provide the lens motion measurement. These factors make it difficult to directly model intrinsics as a function of lens motion, so we have to resort to a hardware-independent approach.

If we know the geometry property of the observed object, we can recover all camera parameters using an estimation method. This is often known as camera calibration. Such methods require ample number of features and are often assisted with carefully-designed calibration patterns~\cite{tsai1987versatile,willson1994modeling,li1996some,heikkila1997four,zhang2000flexible,zhang2004camera,scaramuzza2006toolbox,richardson2013aprilcal} to increase accuracy. Among existing calibration methods, self-calibration~\cite{faugeras1992camera,hartley1994self,hartley1999camera,wildenauer2012robust} (or auto-calibration) does not rely on calibration pattern, and finds the camera intrinsics through projective geometry properties existing in image sequences (e.g. absolute conic~\cite{hartley2003multiple}). Existing calibration methods provide a way to estimate intrinsics, but they cannot be directly applied to our problem because 1) camera intrinsics are dynamic when OIS is activated, and 2) there may not be enough corresponding features in a single frame to recover intrinsics accurately. Therefore, we propose a neural network-based approach to address these problems. After trained in lab settings, DIME-Net is able to capture the dynamic intrinsic properties and infer high quality intrinsics in applications with a small set of features from nature objects.



Although there is no prior work on the dynamic intrinsic estimation directly, the design of DIME-Net is inspired by existing progress in learning-based methods. Recent research begins to transfer the knowledge in geometry domain into the network architecture design for geometry related problem such as pose estimation~\cite{kendall2017geometric,Parameshwara_2022_CVPR} and 3D reconstruction~\cite{laidlow2019deepfusion,Boulch_2022_CVPR_shapenet,ding2022transmvsnet}. In camera pose estimation, PoseNet~\cite{kendall2017geometric} uses CNNs and fully connected layers to solve camera pose regression. These works focus on camera poses which are extrinsics. Although being different from the intrinsic parameter estimation, their methods enforcing the geometric constraints (e.g. reprojection error~\cite{kendall2017geometric}, position and orientation error~\cite{brahmbhatt2018geometry} ) into the loss function for training shed light on how to approach our problem. We employ reprojection error as loss function which is enabled by building on the recent progress on the PnP~\cite{lepetit2009epnp,CampbellAndLiu,chen2020end} problem. BPnP~\cite{chen2020end} considers the optimization as a layer and enables the backpropagation of network as a whole with the help of the implicit theorem. This eventually enables us to employ the reprojection error as our loss function in DIME-Net training design.

In our design, DIME-Net use MLP to approximate the manifold that characterizes dynamic intrinsics. There are existing methods using learning based approach for manifold and distance field approximation~\cite{eccv_2016_image_manifold, mildenhall2020nerf}. Specifically, existing effort has been made on using MLP to represent manifold field. For example, Moser Flow~\cite{rozen2021moser} uses MLP to represent geometry manifold such as Torus. Pose-NDF~\cite{tiwari22posendf} designs a generative model with geometry implicit function representing as feature to create human pose sequence in a manifold.

\section{OIS Effect Mitigation Framework and Problem Definition}

Let us first introduce imaging process and analyze why existing OIS effect mitigation scheme is problematic before introducing our framework and problem definition.

\subsection{Perspective Projection under OIS}

Before we introduce camera imaging, let us define the following coordinate systems and points in them,
\begin{description}
\item[$\{C\}$] 3D camera coordinate system (CCS), where its origin is at the camera center, and its X-axis and Y-axis parallel to the horizontal and vertical axes of its image $2$D coordinate $\{I\}$ respectively.
\item[$\{W\}$] is a fixed 3D world coordinate system.
\item[$\mathbf{x}$] is a homogeneous 3-vector describing a $2$D point position in $\{I\}$, $\mathbf{x}\in\mathbb{P}^{2}$, $2$D projective space.
\item[$\mathbf{X}$] is a 3-vector describing a $3$D point position. As a convention, we use left superscript to indicate the reference frames of $3$D points. For example, ${^{W}\mathbf{X}}$ is a point in $\{W\}$.
\end{description}
All 3D coordinate systems are right-handed system. For a regular camera and according to pinhole perspective projection model, it projects a 3D point ${^{W}\mathbf{X}}$ to a 2D image point $\mathbf{x}$ that can be described by the following model~\cite{hartley2003multiple}
\begin{equation}\label{eq:perspective-project-model}
\mathbf{x} = \lambda \mathrm{K} [^{C}_{W}\mathrm{R}\ ^{C}_{W}\mathbf{t}]\begin{bmatrix}^{W}\mathbf{X} \\ 1\end{bmatrix},
\end{equation}
where $\lambda$ is a scalar and matrix
$\mathrm{K} = \begin{bmatrix}
f_x & 0 & c_x \\
0 & f_y & c_y \\
0 & 0   &  1
\end{bmatrix}$ is the intrinsic matrix of the camera, $f_x$ and $f_y$ are focal lengths in pixel counts using pixel width and height, respectively, and $(c_x,c_y)$ is principal point location on the image. Note that this is a 4-DoF intrinsic matrix model which fits most cameras. We also call $\mathrm{K}$ intrinsics for brevity. Similarly $[^{C}_{W}\mathrm{R},\ ^{C}_{W}\mathbf{t}] \in \mathcal{SE}(3)$, is called extrinsics. If 3D points live in $\{C\}$, then $^{C}_{W}\mathrm{R} = \mathrm{I}_{3}$ becomes an identity matrix and $^{C}_{W}\mathbf{t} = \mathbf{0}_{3}$, a zero vector in $\mathbb{R}^{3}$.

When the camera is equipped with OIS and OIS is activated, both the relative orientation and distance between the lens and the 2D imaging sensor are no longer constants in the camera. From Fig.~\ref{fig::OIS}, each element in $\mathrm{K}$ is a function of lens pose $[\mathrm{R}_{\mbox{\tiny lens}}, \mathbf{t}_{\mbox{\tiny lens}}] \in \mathcal{SE}(3)$. Hence we write it in function format $\mathrm{K}(\mathrm{R}_{\mbox{\tiny lens}}, \mathbf{t}_{\mbox{\tiny lens}})$.

One immediate thought would be if we can directly model function $\mathrm{K}(\mathrm{R}_{\mbox{\tiny lens}}, \mathbf{t}_{\mbox{\tiny lens}})$ based on lens motion $(\mathrm{R}_{\mbox{\tiny lens}}, \mathbf{t}_{\mbox{\tiny lens}})$. Unfortunately, it is very difficult to do so due to lack of information about OIS system design for each individual mobile devices. Depending on how sophisticated the OIS system is, the camera lens may have up to 5 DoFs, although a typical mobile device camera may only has 2 rotational DoFs due to cost and size concerns. Lack of detailed information about motion model is not the only issue. Also, we do not have access to the lens motion feedback since most device software development kits (SDKs) do not provide it. Finally, we do not know which time epoch the OIS aligns frames to. These factors determine that modelling OIS is not a viable approach and we have to opt for a data-driven approach that can be used in a wide range of devices or OIS types.

\subsection{Existing OIS Effect Mitigation Scheme} \label{sc:exising_kc}

The dynamic intrinsics immediately lead to a problem for any vision algorithm that requires constant or known $\mathrm{K}$ matrix. More specifically, any 3D reconstruction or camera pose estimation algorithms would be severely impacted. Existing practices adopted by cellphone manufacturers such as Apple\texttrademark\ or Google\texttrademark\ often resort to a prior approximated camera intrinsic denoted as $\mathrm{K}_{c}$ on reduced resolution images in their SDKs. Such $\mathrm{K}_{c}$ is often obtained by averaging a large number of $\mathrm{K}$'s at different OIS states or at its neutral stationary positions. For $\mathrm{K}_{c}$, we can associate it with extrinsics which defines a unique camera frame $\{C_{0}\}$.

In fact, we also have tested the prior $\mathrm{K}_{c}$ in our experiment setup using PnP problem as an example~\cite{lepetit2009epnp}.
Denote the $i$-th $2$D point as $\mathbf{x}_{i}$. The $2$D and $3$D point correspondences are defined as $\{\mathbf{x}_{i}\leftrightarrow{^{W}\mathbf{X}}_{i}:i=1,\cdots,n\}$, where $n$ is the number of the total point correspondences. PnP algorithm computes camera pose using the 2D-3D correspondences by minimizing reprojection error
\begin{equation}\label{eq:PNP}
    [^{C}_{W}\mathrm{R},\  ^{C}_{W}\mathbf{t}]=\argmin \sum_{i}\left\|\lambda_{i}\mathrm{K}({^{C}_{W}\mathrm{R}{^{W}\mathbf{X}}_{i}}+{^{C}_{W}\mathbf{t}})-\mathbf{x}_{i}\right\|^{2}_{\Sigma},
\end{equation}
where $\|\cdot\|_\Sigma$ is the Mahalanobis norm with the covariance matrix $\Sigma$ for pixel location distribution.

To test the quality of $\mathrm{K}_{c}$, we assume  $\mathrm{K}=\mathrm{K}_{c}$ when solves~\eqref{eq:PNP}. Our point correspondences are from a precise calibration pattern as inputs for the PnP. When the camera resolution is $4032\times3024$, the resulting average reprojection error from PnP is about $3.44$ pixels for an OIS-equipped Samsung Galaxy 8 phone camera whereas that of a camera without OIS can reach $0.42$ pixels under the same settings. In real world applications, the average reprojection error would be much higher because pixelization error from real scene is much higher than the precise and sharp inputs from the calibration pattern. Higher error would cause the algorithm hard to converge under noisy inputs. Consequently, the existing practices are to lower the image resolution to increase the pixel size. This approach is to sacrifice image resolution and camera pose accuracy for algorithm stability, which is not ideal because we cannot fully utilize the true potential of the camera resolution.

\subsection{OIS Intrinsics Rectification Framework} \label{ssc:framework}

One immediate idea is to try to rectify $\mathrm{K}_{c}$. If an accurate $\mathrm{K}$ can somehow be obtained in real time, then the problem is solved. Again, let us use PnP as an example to show how such approach works. It is worth noting that our framework can be easily extended to other applications in 3D scene reconstruction or motion estimation. One quick thought would be if we can add $\mathrm{K}$ as the additional decision variable in the estimation problem in \eqref{eq:PNP} to address the issue. Unfortunately, this would not work because the number of point correspondences in an application is usually insufficient or unevenly distributed which cannot meet the necessary condition to estimate a good quality $\mathrm{K}$.

Since we do not have a clear pathway to estimate $\mathrm{K}$ analytically, the idea becomes if we could find a data-driven approach. The overall framework is illustrated in Fig.~\ref{fig::system_diagram} with three main blocks as follows.

The first step (Box 1(a)) is the initial pose estimation using the prior $\mathrm{K}_{c}$. We know this pose estimation will not be accurate enough, but its residual error are caused by discrepancy between $\mathrm{K}_{c}$ and the actual $\mathrm{K}$ and hence will be important input to next step.

The second step (Box 1(b)) is to recover the $\mathrm{K}$ and the third step (Box 1(c)) is pose refinement with the newly-obtained $\mathrm{K}$ which is simply to re-solve PnP problem with the new $\mathrm{K}$. It is clear that the second step is the key problem here. Let us define this problem,

\begin{Def}
Given $\mathrm{K}_{c}$ and $n$ point correspondences $\{\mathbf{x}_{i}\leftrightarrow{^{C_{0}}\mathbf{X}}_{i}\}^{n}_{i=1}$, design and train DIME-Net to represent $f_{\mathrm{K}}$ manifold that can be used to
predict the dynamic intrinsic camera matrix $\mathrm{K}$.
\end{Def}
Here we assume that nonlinear lens distortion has been removed from images. Cameras with OIS usually have nonlinear lens distortion removed to facilitate OIS.

\begin{figure*}[htb!]
    \centering
    \vspace{0.45in}
    \includegraphics[width=6.5in, trim={1cm 7.5cm 1cm 7.5cm}]{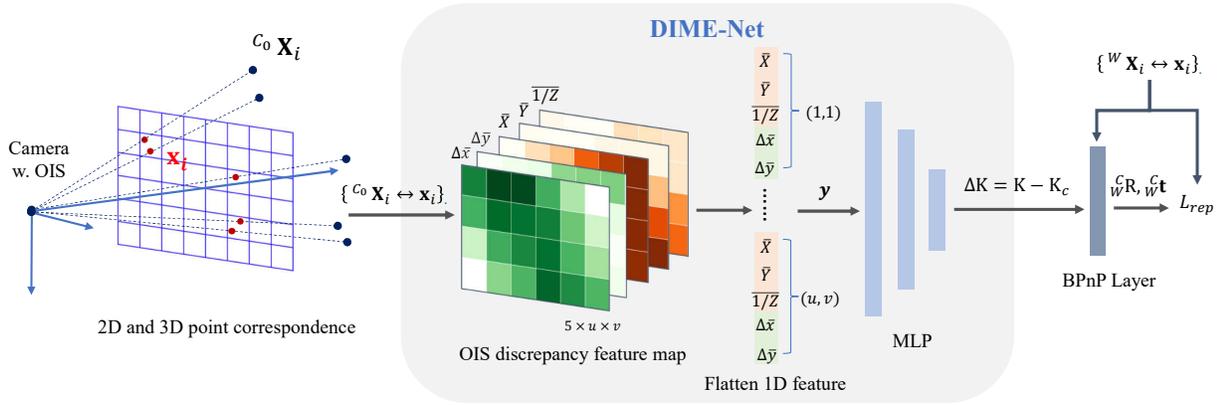}
    \caption{DIME-Net architecture and training scheme. This pipeline reflects the process of training the DIME-Net. In the inference stage, a user only needs the grey box to estimate $\mathrm{K}$ and the pose can be calculated using standard PnP algorithm~\cite{lepetit2009epnp}. }
    \label{fig::network_design}
\end{figure*}

\section{DIME-Net Design and Training}\label{sec::OIS_K_predictor}
The  OIS actuation-caused $\mathrm{K}$ variation can be considered as a $f_{\mathrm{K}}$ manifold despite that we do not have close form representation of  $\mathrm{K}(\mathrm{R}_{\mbox{\tiny lens}}, \mathrm{t}_{\mbox{\tiny lens}})$. In fact, lens pose $[\mathrm{R}_{\mbox{\tiny lens}}, \mathrm{t}_{\mbox{\tiny lens}}]$ is just camera extrinsics $[^{C}_{W}\mathrm{R},\  ^{C}_{W}\mathbf{t}]$ in a different reference system under the actual $\mathrm{K}$. Therefore, we know that these point correspondences have to satisfy ~\eqref{eq:perspective-project-model}. On the other hand, Step 1 of Sec.~\ref{ssc:framework} also produces projected points
\begin{equation}\label{eq:manifold-derivation}
\mathbf{x}_c = \lambda_c \mathrm{K}_c \left[^{C_0}_{W}\mathrm{R}
 \ ^{C_0}_{W}\mathbf{t}\right]\begin{bmatrix}^{W}\mathbf{X} \\ 1\end{bmatrix},
\end{equation}
where corresponding variables with subscription $c$ indicate that they are estimated based on $\mathrm{K}_c$. Define $\Delta \mathbf{x} =  \mathbf{x}_{c} -\mathbf{x}$. We know that
\begin{equation}\label{eq:manifold-derivation}
\Delta \mathbf{x} =
\Bigl\{ \lambda_c \mathrm{K}_c \left[^{C_0}_{W}\mathrm{R}
 \ ^{C_0}_{W}\mathbf{t}\right] -
\lambda \mathrm{K} \left[^{C}_{W}\mathrm{R}
 \ ^{C}_{W}\mathbf{t}\right]
 \Bigr\}
 \begin{bmatrix}^{W}\mathbf{X} \\ 1\end{bmatrix}.
\end{equation}
With the same point correspondences, keep in mind that extrinsics $[^{C_0}_{W}\mathrm{R},\  ^{C_0}_{W}\mathbf{t}]$ and $[^{C}_{W}\mathrm{R},\  ^{C}_{W}\mathbf{t}]$ are functions of corresponding intrinsics $\mathrm{K}_c$ and $\mathrm{K}$, respectively. This means that \eqref{eq:manifold-derivation} defines an input-dependent $f_{\mathrm{K}}$ manifold:
\begin{equation}\label{eq:k_manifold}
f_{\mathrm{K}}(\mathrm{K}, \mathrm{K}_c, \{\Delta  \mathbf{x}_i, \mathbf{X}_i, \forall i\})=0.
\end{equation}
It is not difficult to see that $f_{\mathrm{K}}$ becomes less dependent of individual $\{\mathbf{x}_i, \mathbf{X}_i\}$ as $i$ grows large. At this stage, $f_{\mathrm{K}}$ can be used to predict $\mathrm{K}$ for small number of correspondences. This inspires us to develop a data-driven approach to represent $f_{\mathrm{K}}$ manifold using our DIME-Net. The construction of the approximated $f_{\mathrm{K}}$ manifold vector field, $\mathrm{K}_c$ and $\{\Delta  \mathbf{x}_i, \mathbf{X}_i, \forall i\}$, mapping from input feature vector to the dynamic $\mathrm{K}$ is the DIME-Net training process. It can be done with carefully-collected data under different OIS states with calibration patterns under lab settings. Later in the application, this DIME-Net can be used as $\mathrm{K}$ predictor.

Fig.~\ref{fig::network_design} shows our DIME-Net architecture. We first design the input for DIME-Net which converts the point correspondences and the prior camera matrix into the $1$D OIS discrepancy feature. Given the $1$D OIS discrepancy feature, DIME-Net utilizes the MLP to rectify the camera intrinsics. We will explain how we design DIME-Net with its unique feature, network structure and network loss function.


\subsection{OIS Discrepancy Feature}
\label{sec:feature}

Let us begin with notation definition. Denote the 3D position  ${^{{C}_0}\mathbf{X}}_{i}:=[X_{i}, Y_{i}, Z_{i}]^\mathsf{T}\in\mathbb{R}^{3}$ and the corresponding $2$D pixel position $\tilde{\mathbf{x}}_{i}:=[x_{i}, y_{i}]^\mathsf{T}\in\mathbb{R}^{2}$ where symbol \~\ on a variable means that it is in inhomogeneous coordinate. Note that 3D points are in $\{{C}_0\}$ that the manifold will be defined in $\{{C}_0\}$ instead of $\{W\}$ as \eqref{eq:manifold-derivation}. This change makes the neural network not sensitive to the choice of world coordinate system. There are $3$ steps to obtain the input feature of the neural network: (1) point-based OIS discrepancy feature conversion, (2) grid-based OIS discrepancy feature conversion and (3) $1$D OIS discrepancy feature flattening.

\subsubsection{Point-Based OIS Discrepancy Feature}
Each point-based OIS discrepancy feature is composed by two main components: (1) the inhomogeneous representation of $\Delta \mathbf{x}$ which is named as projection model discrepancy (PMD) feature because it is the 2D reprojection error between the observed image points and their reprojected points using $\mathrm{K}_{c}$, and (2) $3$D point position in $\{C_{0}\}$. The PMD is the direct result of $\mathrm{K}$ change introduced by OIS~\cite{yeh-OIS-detect-case-2022} when $3$D point position in $\{C_{0}\}$ are given.


Denote the PMD of $\mathbf{x}_{i}\leftrightarrow{^{C_0}\mathbf{X}}_{i}$ as $\begin{bmatrix}\Delta x_{i}, \Delta y_{i}\end{bmatrix}^\mathsf{T}\in\mathbb{R}^{2}$, and we have
\begin{equation}\label{eq::motion_discrepancy}
    \begin{bmatrix}\Delta x_{i} \\ \Delta y_{i}\end{bmatrix}=\begin{bmatrix}\mathbf{k}^{1}_{c}\\ \mathbf{k}^{2}_{c}\end{bmatrix}\begin{bmatrix}X_{i}/Z_{i}\\ Y_{i}/Z_{i} \\ 1\end{bmatrix} - \begin{bmatrix}x_{i}\\ y_{i}\end{bmatrix},
\end{equation}
where $\mathbf{k}^{j}_{c}$ is the $j$-th row of $\mathrm{K}_{c}$. It is worth noting that \eqref{eq::motion_discrepancy} is the simplification of \eqref{eq:manifold-derivation} in $\{C_0\}$. We then concatenate PMD and the $3$D point position in $\{C_{0}\}$ to form the point-based OIS discrepancy feature. Denote the point-based OIS discrepancy feature of $\mathbf{x}_{i}\leftrightarrow{^{C_{0}}\mathbf{X}}_{i}$ as $\mathbf{f}_{i}$ which is defined as
\begin{equation}\label{eq::point_based_OIS_feat}
    \mathbf{f}_{i} := \begin{bmatrix}\Delta x_{i}, \Delta y_{i}, X_{i}, Y_{i}, 1/Z_{i}\end{bmatrix}^\mathsf{T}\in\mathbb{R}^{5}.
\end{equation}
It is worth noting that here we employ the inverse depth $1/Z_{i}$ since the inverse depth is linear to $\mathbf{k}^{1}_{c}$ and $\mathbf{k}^{2}_{c}$ in \eqref{eq::motion_discrepancy} which makes the model more linear and fits  better to the neural network.

\subsubsection{Grid-Based OIS Discrepancy Feature}
However, there are two remaining issues when using the point-based features: 1) the point-based feature number is not fixed because it is input-dependent, and 2) the order of features should be irrelevant. In fact, the features should be related to 2D position in the image. If we blindly feed the point-based features into a neural network, we would run into issues because 1) the neural network would need a fixed input dimension, and 2) the neural network would inevitable learn the order of inputs instead of the spatial location in image. To address these issues, we convert the point-based features into grid-based features by using grid cells and merging feature information within each grid cell. This approach fixes the input dimension and order issues since there is a constant number of grid cells and we can arrange the grid feature using the lexicographic order of cells.

First, we create a 2D feature map which has same size as the original image but with 5 channels that contains point-based OIS discrepancy feature $\mathbf{f}_{i}$ for each point correspondence. The feature is indexed by the its 2D point position $(x_i,y_i)$ in the 2D feature map. Next we divide the 2D point based feature map into a $2$D grid pattern consisting of $u\times v$ equal-sized square grid cells. For each cell, we average the point-based features in the cell to be the corresponding grid-based feature. Let $\begin{bmatrix}a_{k},b_{j}\end{bmatrix}^\mathsf{T}$ be the bottom-right corner point pixel position of the grid in the $j$-th row and the $k$-th column of the $2$D grid pattern. The point-based feature set residing in the grid in the $j$-th row and the $k$-th column is defined as
\begin{equation}
    \mathcal{F}_{j,k}:=\Big\{ \mathbf{f}_{i}:x_{i}\in[a_{k-1},a_{k})\ \text{and}\ y_{i}\in[b_{j-1},b_{j}) \Big\}.
\end{equation}

Denote the grid-based OIS discrepancy feature of the grid in the $j$-th row and the $k$-th column as $\mathbf{y}_{j,k}$. The grid-based OIS discrepancy feature $\mathbf{y}_{j,k}$ is defined as
\begin{equation} \label{eq::img_patch_OIS_feature}
\begin{aligned}
    \mathbf{y}_{j,k}:=&\frac{1}{|\mathcal{F}_{j,k}|}\sum_{\mathbf{f}_{i}\in\mathcal{F}_{j,k}}\mathbf{f}_{i}\\
    =&\begin{bmatrix}\Delta \overline{x}_{j,k}, \Delta \overline{y}_{j,k},\overline{X}_{j,k},\overline{Y}_{j,k},\overline{1/Z}_{j,k}\end{bmatrix}^\mathsf{T}\in\mathbb{R}^{5},
\end{aligned}
\end{equation}
where $|\cdot|$ is the set cardinality and symbol $^{-}$ on a variable indicates the average value. 

\subsubsection{$1$D OIS Discrepancy Feature Flattening}
We flatten the grid-based features to be one dimension as the final input for the MLP in the next step. Denote the flattened feature vector as $\mathbf{y}$. The flattened vector $\mathbf{y}$ can be obtained by concatenating the grid-based features,
\begin{equation}\label{eq:1Dfeature}
    \mathbf{y}:=\left[\mathbf{y}^\mathsf{T}_{1,1}, \mathbf{y}^\mathsf{T}_{1,2},  \ \dots,\  \mathbf{y}^\mathsf{T}_{u,v}\right]^\mathsf{T} \in\mathbb{R}^{m_{y}},
\end{equation}
where $u$ and $v$ are the numbers of the grid cells in row and column, respectively, and dimension $m_{y}=5\cdot u\cdot v$.

\subsection{DIME-Net Architecture and Loss Function}


As illustrated in Fig.~\ref{fig::network_design}, in order to learn the $f_{\mathrm{K}}$ manifold, we design a generative model using the multi-layer perceptron to generate the dynamic $\mathrm{K}$ from $\mathbf{y}$. We employ the geometric error as a loss function to link the network's performance to camera projection model.

\subsubsection{Multi-layer Perceptron}
As introduced in previous sections, $\mathbf{y}$ is the feature vector of point correspondence that describe camera OIS effect. Our goal is to generate the camera intrinsic $\mathrm{K}$ from $\mathbf{y}$. From OIS feature perspective, intrinsics $\mathrm{K}$ is a latent variable that directly describes the camera model. We design an MLP to be an antoencoder-style mapping from a high dimension feature variable $\mathbf{y}$ to a low dimension latent variable $\mathrm{K}$. Specifically, we employ a fully connected 3-layer perceptron to generate $\mathrm{K}$. We design the network output to be $\Delta \mathrm{K} = \mathrm{K} - \mathrm{K}_c$. The output layer has 4 nodes that represent four components of $\Delta \mathrm{K}$: $\Delta f_{x}$, $ \Delta f_{y}$, $ \Delta c_{x}$ and $\Delta  c_{y}$. This design helps regulate the network. In special case when input vector $\mathbf{y}=\mathbf{0}_{m_{y}}$, the network should output $\mathbf{0}_{4}$ so that $\Delta\mathrm{K} = \mathrm{0}_{3\times 3}$ and $\mathrm{K}=\mathrm{K}_c$ due to lack of information, which also ensures the network stability.


\subsubsection{BPnP Layer in Training and Loss Function Design}
For the network training, we employ reprojection error as the loss function. This directly ties network performance with model quality. Given the predicted intrinsics $\mathrm{K}$ and extrinsics ${^{C}_{W}\mathrm{R}}$ and ${^{C}_{W}\mathbf{t}}$ and the point correspondences $\{\mathbf{x}_{i}\leftrightarrow{^{W}\mathbf{X}}_{i}\}$, the loss can be calculated by
\begin{equation}\label{eq::loss}
    L_{\mbox{\footnotesize rep}} = \sum_{i}\bigl\|\lambda_{i}\mathrm{K}({^{C}_{W}\mathrm{R}{^{W}\mathbf{X}}_{i}}+{^{C}_{W}\mathbf{t}})-\mathbf{x}_{i}\bigr\|^{2}_{\Sigma}.
\end{equation}
Note that we would need extrinsics $[{^{C}_{W}\mathrm{R}},{^{C}_{W}\mathbf{t}}]$ to compute the loss function. To obtain extrinsics and enable the network end-to-end training, we connect the network with a BPnP layer~\cite{chen2020end} to estimate $[{^{C}_{W}\mathrm{R}}, {^{C}_{W}\mathbf{t}}]$ from the predicted $\mathrm{K}$. Compared with the general PnP solver, BPnP considers the optimization as a layer and enables the backpropagation of network as a whole by the help of the implicit theorem~\cite{krantz2002implicit}. Using reprojection error~\cite{hartley2003multiple} as our loss function makes the overall model like a maximum likelihood estimator. Common loss functions like L1 or L2 norm are algebraic distance which is not robust and can lead to a spurious solution since it does not contain the geometric meaning. Reprojection error, on the other hand, is a geometric distance. Hence, the loss function in \eqref{eq::loss} can guide the network in learning the $f_{\mathrm{K}}$ manifold.


\subsection{Training and Inference Using DIME-Net}\label{sec::train&inf}

\begin{figure}[htbp]
\centering
\subfloat[\label{fig::train_sample}]{\includegraphics[ width=0.2\textwidth]{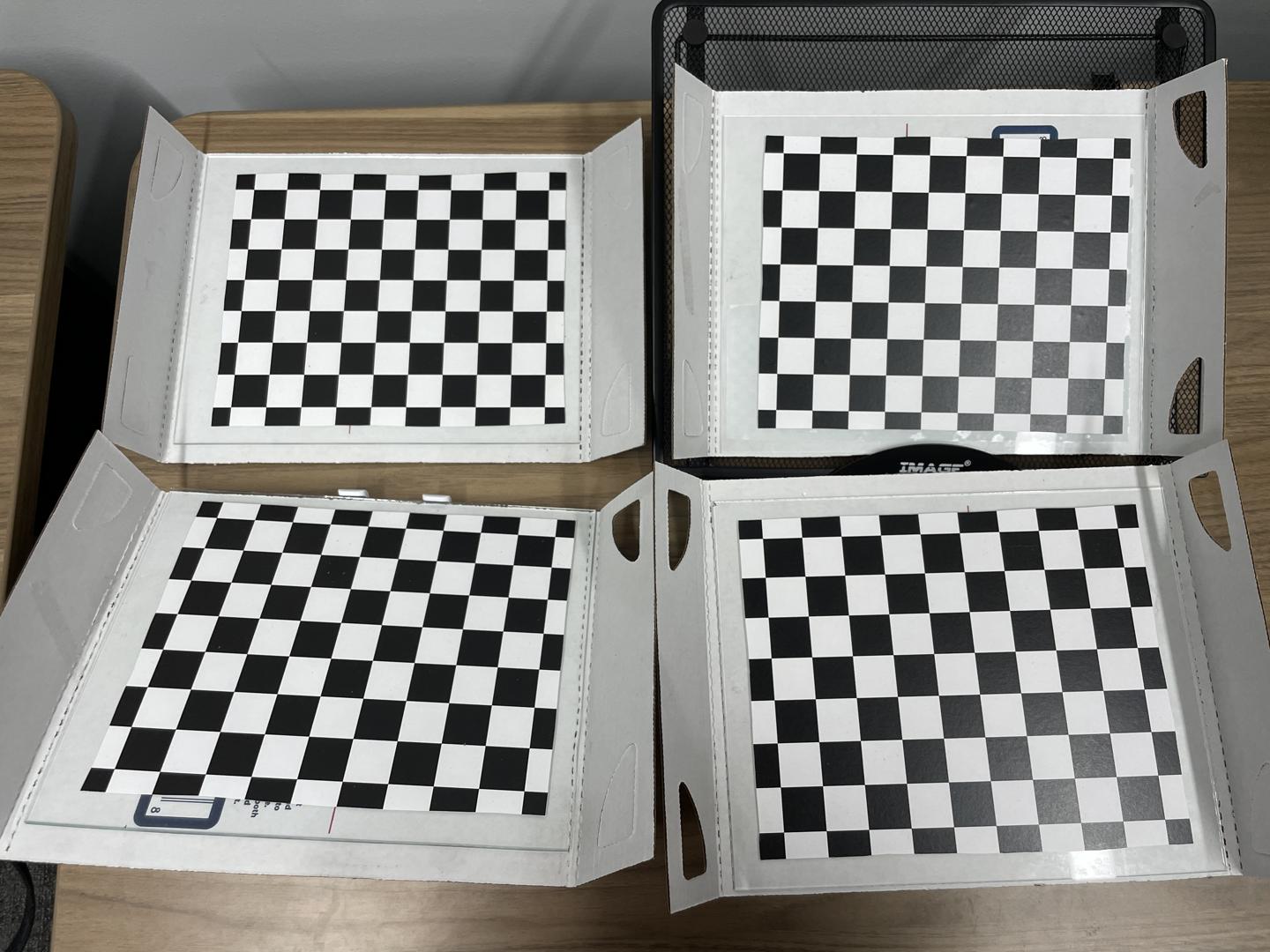}}\hspace*{.2in}
\subfloat[\label{fig::lego_sample}] {\includegraphics[ width=0.2\textwidth]{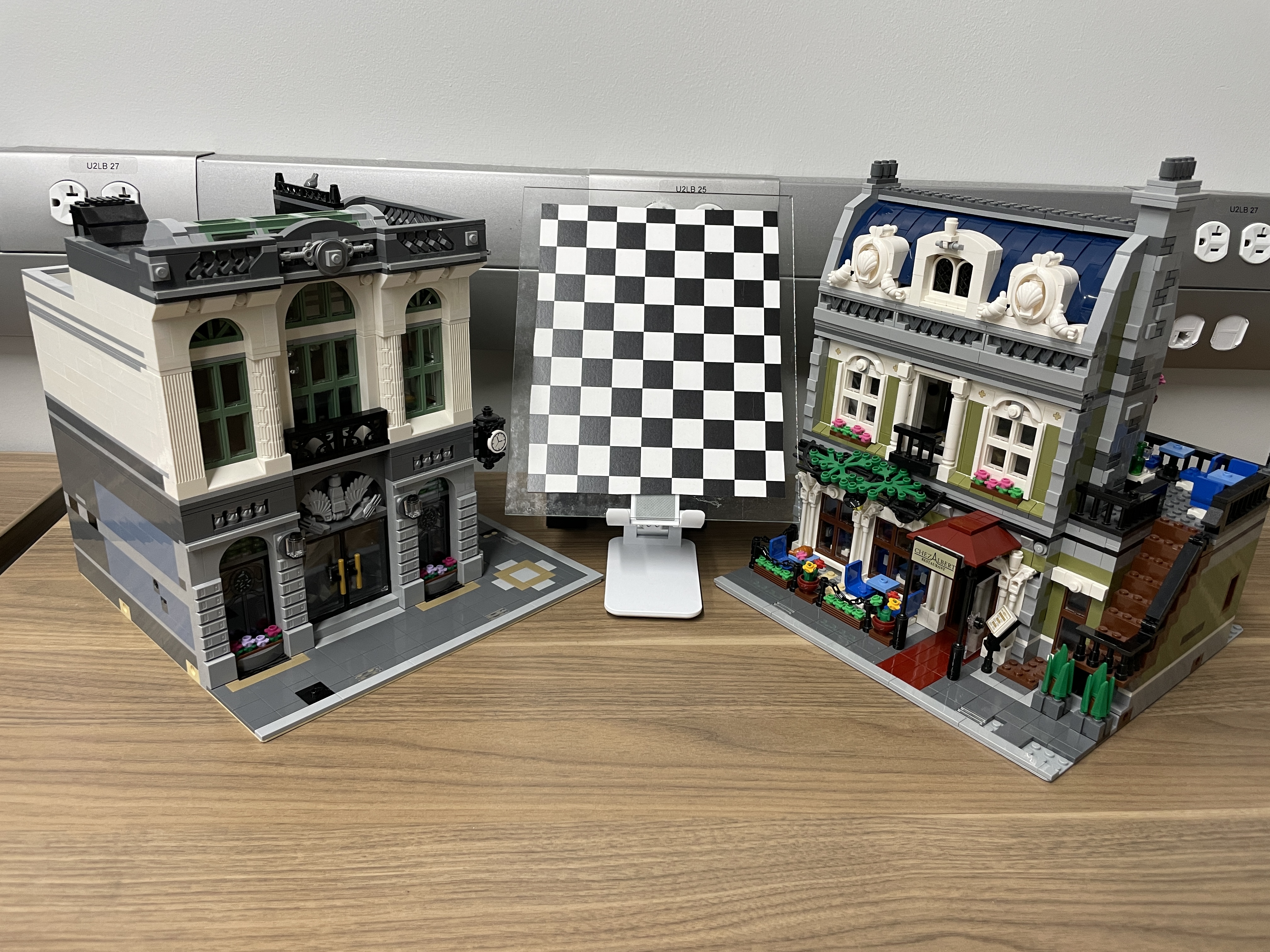}}
\caption{(a) Example image of the training inputs for DIME-Net using the calibration rig. (b) Example image of natural object feature test setup where two LEGO buildings are the natural objects. }
\label{fig::train_infer_setup}
\end{figure}

To gather good training samples, as shown in Fig.~\ref{fig::train_sample}, we have designed a calibration rig. It contains 4 checkerboard pattern located at 4 different planes. Each checkerboard pattern contains $8\times10$ inner vertices and is deployed on a planar glass ensure flatness. Each cell side length is $22.0$ mm. 3D points positions are computed in $\{C_{0}\}$ and 2D points are readouts from vertex coordinates in the image. It is worth noting that the 4-checkerboard rig design allows us to directly obtain $\mathrm{K}$ for each image using calibration procedure because there are enough inputs to estimate both $\mathrm{K}$ and extrinsics. This is very important in training and verification because it provides ground truth. With this setup, we can obtain a set of 2D-3D correspondences with a moving camera at different perspectives that covers the normal working range of the camera.

We use the accurate point correspondences to obtain the feature vector $\mathbf{y}$ for the network training by monitoring the convergence of the loss function in \eqref{eq::loss}. The good coverage of the training data ensures that our neural network can approximate $f_{\mathrm{K}}$ manifold with good accuracy.

With a trained network, we can deploy it for inference in applications. Our DIME-Net has the ability to predict $\Delta \mathrm{K}$ given the input feature vector in \eqref{eq:1Dfeature} converted from the 2D-3D point correspondence set.




\section{Experiments}
We have implemented our DIME-Net using PyTorch~\cite{paszke2019pytorch}.  We first perform an ablation study of our DIME-Net. Then we evaluate the inference accuracy of our DIME-Net using both calibration rig and nature object features. Let us introduce our OIS datasets.

\subsection{Calibration Rig OIS Datasets}\label{sec::calib_rig_OIS_dataset}
We have collected data under OIS effect using our calibration rig detailed in Sec.~\ref{sec::train&inf}. To activate OIS, we hand-hold the camera and capture images with different poses. We use three different cameras as detailed in Tab.~\ref{tab:CB}. For each camera, we collect a dataset and split it into training set and testing set (shown as number of images in ``Train'' and ``Test'' columns).

For each device, we also obtain  $\mathrm{K}_{c}$ according the method in Sec.~\ref{ssc:framework}.
In addition, for each image, we also use the 4-board as input to estimate $\mathrm{K}^{\star}$ using camera calibration method. The calibration process method yields reprojection error $e^{\star}$. The average reprojection errors are shown in the Avg($e^{\star}$) column with unit as pixels which provide a baseline for the best possible performance for reprojection error.

\begin{table}[ht!]
    \centering
    \caption{Calibration rig OIS image  datasets}\label{tab:CB}
    \begin{tabular}{l | c c c c} \toprule
        Device & Resolution & Train & Test &  Avg($e^{\star}$) \\ \hline
        Samsung Galaxy S8 & $4032\times3024$ & 185 & 47 & 0.45  \\
        iPhone 12 Pro & $4032\times3024$ & 164 & 42 & 0.46\\
        iPad mini 6 & $4032\times3024$ & 224 & 57 & 0.32 \\  \bottomrule
    \end{tabular}

\end{table}

\subsection{DIME-Net Ablation Study}
Here we test the impact of different feature setups for DIME-Net performance using Samsung Galaxy S8 data from Tab.~\ref{tab:CB}.

\subsubsection{$2$D Grid Resolution and Occupancy Tests}
Now we test how the $2$D grid resolution and grid cell occupancy can affect the DIME-Net performance under reprojection error $e$. Define the average reprojection error as Avg($e$) which is used as a primary metric. The $2$D grid resolution $u\times v$ determines the number of  point-based OIS discrepancy features in each cell and affects the uncertainty of the grid-based OIS discrepancy features which are the direct inputs of the DIME-Net. The $2$D grid occupancy, on the other hand, indicates the distribution of the OIS information preserved. As shown in Tab.~\ref{table::image_patch_sensitivity}, we have chose $3$ different $2$D grid resolution: $16\times12$, $12\times9$ and $8\times6$. To simulate the occupancy, we uniformly sample the cells and empty the $2$D and $3$D point correspondences in the cells. The ratio of the emptied cell is measured by
$\eta = 1 - \frac{m'_{p}}{m_{p}},$
where $m_{p}$ and $m'_{p}$ are the number of cells with non-zero OIS features before and after the sampling, respectively. The $2$D grid occupancy then is measured by $ \gamma = \frac{m_{p}}{u\cdot v}.$


\begin{table}[htbp]
	\centering
    \caption{Avg($e$) in pixels vs. $2$D grid resolution and occupancy in different $\eta$ and $\gamma$. Smaller is better.  Best results are in boldface.}
	\label{table::image_patch_sensitivity}
	\begin{tabular}{c | c c| c c| c c }
		\toprule[0.8pt]
        & \multicolumn{6}{ c }{$2$D grid resolution}\\ \cline{2-7}
        & \multicolumn{2}{ c |}{$16\times12$} & \multicolumn{2}{ c| }{$12\times9$} & \multicolumn{2}{ c }{$8\times6$} \\
        \cline{2-3} \cline{4-5} \cline{6-7}
        $\eta\%$& $\gamma\%$ & Avg($e$) & $\gamma\%$ & Avg($e$) & $\gamma\%$ & Avg($e$)\\ \hline
0 &  64.5 & 0.73  & 78.6 & 0.75 & 96.4 & \textbf{0.68} \\
20  & 52.1 & 0.87 & 62.6 & 0.97 & 79.3 & 0.92   \\
40  & 38.9 & 1.29 & 46.0 & 1.33 & 57.7 & 1.45  \\
60  & 26.6 & 1.77 & 30.5 & 1.88 & 36.6 & 2.05  \\
80  & 13.4 & 2.47 & 14.8 & 2.49 & 19.7 & 2.52  \\
     \hline
	\end{tabular}
	\centering
\end{table}

Tab.~\ref{table::image_patch_sensitivity} shows that the $2$D grid resolution with $8\times6$ can achieve the lowest Avg($e$). It is expected since the size of the cell and the number of the point-based OIS discrepancy features increases as grid resolution reduces. The average in \eqref{eq::img_patch_OIS_feature} reduces feature noise when there are more point features in each cell. The lowest Avg($e$) of $0.68$ pixels is close to the calibration accuracy of $\mbox{Avg}(e^{\star})=0.45$ in Tab.~\ref{tab:CB} which confirms that our DIME-Net works effectively in learning the $f_{\mathrm{K}}$ manifold.
The results show the effective design of the DIME-Net feature because it is capable of predicting intrinsics even when the grid occupancy is extremely low.

\subsubsection{OIS Discrepancy Feature Tests}
Next, we examine OIS discrepancy feature components in \eqref{eq::motion_discrepancy}. Again, Avg($e$) in pixels is used as the metric. We choose $8\times6$ for the $2$D grid resolution since it has the best performance. We compare five different setups.

\begin{itemize}
   \item[A.\ ] Complete OIS discrepancy feature in \eqref{eq::img_patch_OIS_feature} using both PMD and $3$D point positions. 
   \item[B.\ ] Only use PMD in \eqref{eq::motion_discrepancy}. 
   \item[C.\ ] Combine PMD with inverse depth $1/Z$. 
   \item[D. ] Similar to ``C", but we combine PMD with $X$ and $Y$ positions of $3$D points. 
   \item[E. ] Only use $3$D point positions. 
\end{itemize}

Tab.~\ref{tab:feature compare} shows that option A achieves the lowest Avg($e$) which means that all features are necessary to achieve the best result. This is not surprising since \eqref{eq:manifold-derivation} has told us that. What is interesting is that the performance of options B-D is slightly worse than that of A, which indicates that PMD is the dominating feature. 

\begin{table}[ht!]
	\centering
    \caption{Avg($e$) comparison of different  feature combination}\label{tab:feature compare}
    \begin{tabular}{c c c c c c }
        \toprule
        & A    & B    & C & D & E \\ \cline{2-6}
        Avg$(e)$ & \textbf{0.68} & 0.78   & 0.78
  & 0.75 & 2.58\\ \bottomrule
    \end{tabular}
\end{table}



\subsection{Inference Accuracy Comparison}
After knowing the best setup for DIME-Net, we are ready to compare it to the state-of-the-art in inference test. 

\textbf{Evaluation Metric for Accuracy:}
We use Avg($e$) as basic performance metric. From Sec.~\ref{sc:exising_kc}, we know the popular existing approach is to employ the prior $\mathrm{K}_{c}$ which is obtained when camera is at the stationary or by averaging a large number of $\mathrm{K}$'s under different OIS states. Let us define Avg($e_{c}$) as its average reprojection error when only using  $\mathrm{K}_{c}$.

We also set up the baseline for comparison. The baseline is characterized by $\mathrm{K}^{\star}$ which is the best intrinsics that one can obtain for the test case. Recall that Avg($e^{\star}$) is its reprojection error. Avg($e^{\star}$) reflects noises in points which is the level of noise that cannot be canceled by adjusting intrinsics without over-fitting. It is not difficult to see that Avg($e^{\star}$) $\leq$ Avg($e_{c}$) given a reasonable large population of point correspondences. It is also clear that if our design is effective, then Avg($e$) should fall between the two. The closer Avg($e$) is to Avg($e^{\star}$), the better it is. This can be measured by a new metric: the average reprojection error reduction ratio,
\begin{equation}
    \rho = \frac{\text{Avg}(e_{c})-\text{Avg}(e)}{\text{Avg}(e_{c})-\text{Avg}(e^{\star})}.
\end{equation}
Higher $\rho$ is more desirable, and $0\leq\rho\leq1$. Now we are ready to compare the inference quality under different datasets.

\subsubsection{Calibration Rig Inference Accuracy Tests}
The first test is done with data shown in Tab.~\ref{tab:CB} based on the calibration rig data.
\paragraph{Point Dropping and Noise Injection Tests}\label{sec::point_drop_noise_inject_tests}
We want to test inference accuracy of DIME-Net after we decrease the number of the point correspondences and/or inject noises to 2D and 3D points. This is important because real world applications do not always have ample amount of features at calibration board point accuracy.
To generate the testing condition of decreased point numbers, we uniformly sample the point correspondences to be dropped. For noise injection, we inject random zero mean Gaussian noise with standard deviation $\sigma^{+}_{x}$ and $\sigma^{+}_{X}$ to the $2$D point and $3$D point, respectively.
It is worth noting that the injected noise $\sigma^{+}_{x}$ and $\sigma^{+}_{X}$ are the additional noise added on the checkerboard vertices. The units of $\sigma^{+}_{x}$ and $\sigma^{+}_{X}$ are pixel and mm, respectively. In this test, we use the Samsung Galaxy S8 camera with $8\times6$ $2$D grid resolution for the DIME-Net.

Tab.~\ref{tab:zero_2d_3d_added_noise} shows the results. Note that in our experimental setup $1$ mm means about $5.45$ pixels (px). The upper half of the table are the results when zero injected noise is added ($\sigma^{+}_{x}$=0 \mbox{px} , $\sigma^{+}_{X}$=0 mm), and the lower half of the table are results when $\sigma^{+}_{x}=3 \ \mbox{px}$ and $\sigma^{+}_{X}=0.1$ mm.  The average reprojection error reduction rate, $\rho$, shows that our DIME-Net is insensitive to the injected noise and the low number of the point correspondences. It remains to be close to or above 90\% until the sample sizes drop to 64 in either cases. Our design has been shown to be effective and robust against the point dropping and noisy inputs.

\begin{table}[ht!]
    \centering
    \caption{Inference accuracy of ($\sigma^{+}_{x}=0\ \mbox{px}, \ \sigma^{+}_{X}=0$ mm) (upper half) and ($\sigma^{+}_{x}=3\ \mbox{px}, \ \sigma^{+}_{X}=0.1$ mm) (lower half) under different sample sizes.}\label{tab:zero_2d_3d_added_noise}
\begin{tabular}{c | c c c c}
\midrule
\#Samples & Avg($e_c$) & Avg($e$) & Avg($e^{\star}$) & $\rho$(\%)  \\  \hline
320  & 3.44 & 0.68 & 0.45 &  92.2 \\
256  & 3.34 & 0.67 & 0.45 &  92.5 \\
192  & 3.43 & 0.69 & 0.45 &  92.2 \\
128  & 3.45 & 0.85 & 0.46 &  87.0 \\
64   & 3.40 & 1.35 & 0.45 &  69.8 \\  \hline\hline
320  & 5.28 & 3.90 & 3.86 &  97.2\\
256  & 5.27 & 3.89 & 3.82 &  95.0\\
192  & 5.28 & 3.88 & 3.83 &  96.8\\
128  & 5.31 & 3.84 & 3.81 &  97.7\\
64   & 5.56 & 4.01 & 3.81 &  88.2\\
\hline
\end{tabular}

\end{table}

\paragraph{Multi-device Tests} We repeat the tests for all three devices using data in Tab.~\ref{tab:CB} under the same settings as upper half of Tab.~\ref{tab:zero_2d_3d_added_noise}. The results in upper half of Tab.~\ref{tab:multi-device} are consistent with previous tests: our DIME-Net achieve over 91\% in $\rho$ in all cases.

\begin{table}[ht!]
    \centering
    \caption{Inference accuracy for three devices and natural objects.}\label{tab:multi-device}
    \begin{tabular}{l| c c c c}
    \toprule[0.8pt]
        Device &Avg($e_c$) & Avg($e$) & Avg($e^{\star}$) & $\rho$(\%) \\ \hline
        Samsung Galaxy S8 &  3.44 & 0.68 & 0.45 & 91.7\\ 
        iPhone 12 pro     & 2.36  & 0.61 & 0.46 & 92.3\\ 
        iPad mini 6       & 1.61  & 0.36 & 0.32 & 96.8\\
        \hline\hline
         Samsung Galaxy S8 (N) & 4.26 & 3.74 & 3.45 & 64.0\\
         \hline
    \end{tabular}
\end{table}

\subsubsection{Nature Object Inference Accuracy Tests}
We also perform inference accuracy comparison test on nature objects since this indicates its performance in real world scenarios where points are not from precise calibration pattern. Here we employ the LEGO buildings as natural objects (See Fig.~\ref{fig::lego_sample}). The $2$D grid resolution is $8\times6$.

It is worth noting that we only rely on points from the LEGO buildings for $\mathrm{K}$ inference. PnP is then computed use both points on LEGO and checkerboard just for evaluation purpose. This allow us to tell if $\mathrm{K}$ is rectified properly by the reprojection error instead of affecting by the in-precise PnP estimation due to unevenly distributed features. Here we estimate $\mathrm{K}^{\star}$ by employing the Maximum Likelihood Estimation (MLE) to refine $\mathrm{K}$ and extrinsics simultaneously when minimizing the reprojection error over all points.

\textbf{LEGO OIS Dataset:}
We employ two LEGO buildings as the natural objects. We manually select/label the features of the LEGO buildings, where there are $47$ features on the left side and $19$ features on the right side in Fig.~\ref{fig::lego_sample}. The corresponding 3D points are obtained from their CAD models and multi-stage alignments with the assistance from the middle checkerboard. We have collected $34$ images with different poses by hand-held camera to activate OIS during the image capture.



The last row of Tab.~\ref{tab:multi-device} shows the results. DIME-Net has achieved over 64\% reduction in reprojection error and proved that it is effective. Moreover, it is worth noting that the results are consistent with that in Tab.~\ref{tab:zero_2d_3d_added_noise} because we only have 66 features for inference and relatively large noise with Avg($e^{\star}$) being 3.45 px.
In fact, the 64\% reduction by using only the natural object features also shows that our training scheme design using the calibration rig successfully learns the $f_{\mathrm{K}}$ manifold since the setups between the LEGO OIS dataset and the calibration rig OIS datasets are different.

\section{Conclusions and Future Work}

To deal with the camera intrinsics variation caused by OIS system, we presented our new DIME-Net, a multi-layer perceptron network designed to rectify camera intrinsics in real time. Using PnP problem as an example, we analyzed OIS system and proposed to use a gridified PMD feature set along with 3D point positions to train DIME-Net using calibration patterns. The trained network became an approximation of the intrinsics manifold that can predict rectified intrinsics in application. We have implemented and extensively tested our design. The experimental results confirmed that our design was robust and effective and can significantly reduce reprojection error. In the future, we will improve our design with better geometry insights and external sensors to future reduce the reliance on number of features required.

{\small
\section*{Acknowledgment}
We thank D. Shell, Y. Xu and Z. Shaghaghian for their insightful discussions. We are also grateful to A. Kingery, F. Guo, C. Qian, and Y. Jiang for their inputs and feedback.

\bibliographystyle{IEEEtran}
\bibliography{Yeh,dez,syxie_ois}
}

\end{document}